\documentclass[a4paper, 10pt, conference]{ieeeconf}      
\usepackage[bookmarks=false]{hyperref}
\usepackage{FG2017}
\usepackage{comment}
\usepackage{hyperref}
\usepackage{epsfig,graphicx,lscape}
\usepackage{placeins}
\usepackage{float}
\usepackage{amsmath}

\FGfinalcopy 

\IEEEoverridecommandlockouts                              
\title{\LARGE \bf
Clothing and People - A Social Signal Processing Perspective
}

\author{\parbox{16cm}{\centering
    {\large Maedeh Aghaei$^1$, Federico Parezzan$^3$, Mariella Dimiccoli $^1$, Petia Radeva$^{1,2}$, Marco Cristani$^3$}\\
    {\normalsize
    $^1$ Faculty of Mathematics and Computer Science, University of Barcelona, Barcelona, Spain\\
    $^2$ Computer Vision Center, Barcelona, Spain\\
    $^3$ Department of Computer Science, University of Verona, Verona, Italy}}
    \thanks{This work was partially founded by projects TIN2012-38187-
C03-01 and SGR 1219 and joint project 2016 grant program, University of Verona. M.A. is supported by an APIF grant from the University of Barcelona. P.R. is partly supported by an ICREA Academia grant.}
}

\begin{document}

\IEEEoverridecommandlockouts\pubid{\makebox[\columnwidth]{978-1-5090-4023-0/17/\$31.00~\copyright{}2017 IEEE \hfill} \hspace{\columnsep}\makebox[\columnwidth]{ }}

\ifFGfinal
\thispagestyle{empty}
\pagestyle{empty}
\else
\author{Anonymous FG 2017 submission\\-- DO NOT DISTRIBUTE --\\}
\pagestyle{plain}
\fi
\maketitle

\begin{abstract}
In our society and century, clothing is not anymore used only as a means for body protection. Our paper builds upon the evidence, studied within the social sciences, that clothing brings a clear communicative message in terms of social signals, influencing the impression and behaviour of others towards a person. In fact, clothing correlates with personality traits, both in terms of self-assessment and assessments that unacquainted people give to an individual. The consequences of these facts are important: the influence of clothing on the decision making of individuals has been investigated in the literature, showing that it represents a discriminative factor to differentiate among diverse groups of people. Unfortunately, this has been observed after cumbersome and expensive manual annotations, on very restricted populations, limiting the scope of the resulting claims. With this position paper, we want to sketch the main steps of the very first systematic analysis, driven by social signal processing techniques, of the relationship between clothing and social signals, both sent and perceived. Thanks to human parsing technologies, which exhibit high robustness owing to deep learning architectures, we are now capable to isolate visual patterns characterising a large types of garments. These algorithms will be used to capture statistical relations on a large corpus of evidence to confirm the sociological findings and to go beyond the state of the art.
\end{abstract}

\section{INTRODUCTION}
Despite rich body of work on the role of behavioral cues in non-verbal social signal processing~\cite{vinciarelli2015new}, deeper understanding of social signals requires further cues discovery. In this respect, some visual behavioral cues such as gesture, posture, gaze, physical appearance, and proxemics have received high attention, while other possible features such as clothing have been traditionally little studied \cite{vinciarelli2009social}.

This is an important lack in the social signal processing literature, since clothing affects behavioral responses in the form of impression formation or person self-perception. Several past studies in the social sciences aimed to assess this influence, showing that formality of the clothing influences impression of others towards a person \cite{fortenberry1978mode} as well as the self-perception of people towards themselves \cite{hannover2002clothing, adomaitis2005casual}. More recently, the influence of clothing on the decision making of individuals has been investigated \cite{stephan2011effects}. 
A few studies also have shown that clothing may be an indicator of ethnicity, culture, socioeconomic status \cite{sybers1962sociological,kerr1990dress}, and even surroundings of the people \cite{murillo2012urban}.

Other studies shows that clothing correlates with the personality traits of people in a way that people with formal clothing perceive actions and objects, the inter-relationship and the intra-relationship between them in a more meaningful manner \cite{slepian2015cognitive}. Clothing can make a person feel comfortable or not in a social situation \cite{hogan3secret} and can be considered as a determinant of how long it takes for strangers to trust one and how much they may trust them \cite{timming2016trust}.  Aforementioned studies are evidences of the importance of clothing in social signaling. Arguably, clothing can be considered as the most evident blueprint of individuals, which is completely dependent on their conscious choices, is not as transient as a gesture, and is more evident than any micro-signals such as a sarcastic smile among the facial expressions.

Various experiments have been previously performed to measure the clothing effect on human behavior. According to the critical review of Johnson \emph{et al.} \cite{johnson2008dress}, the effect of clothing on human behavior, usually is measured in combination with other variables. However, despite the rich body of work, to the best of our knowledge, all of the previous experimental studies were performed and analyzed manually. 

In this position paper, we will sketch the future steps of the first systematic study on which social signals are conveyed by clothing, proposing a framework within the scope of computer vision to measure the clothing effect on the impression that we have on ourselves and that we trigger in the others. More precisely, in a first phase we will investigate the basic visual cues that could be associated to social signals; for example, checking how much tight shirts are associated to the social signal of attractiveness. In a second phase, we will perform a higher level analysis, investigating the types of behaviors that may result from a social interaction, in dependence on the type of garments worn by the interacting people; for example, analyzing interactions between formally versus casually suited individuals. All of this would be possible since computer vision technologies are now mature for a fine-grained analysis of the clothing, providing precise dense segmentations of outfits as results of human parsing algorithms, and automatically recognizing  diverse clothing items \cite{vittayakorn2015runway,yamaguchi2015retrieving,dong2016parsing} and styles~\cite{kiapour2014hipster}.

The paper is organized as follows: in Sec.~\ref{sec:Semiotics} the 
literature on clothing in terms of social sciences is reviewed; in Sec.~\ref{sec:CV}, clothing analysis in terms of human parsing approaches is reported. The core of the paper is Sec.~\ref{sec:Our}, where we discuss our ideas related to the study of clothing under the social signal processing umbrella. The paper ends with some final remarks in Sec.~\ref{sec:Conc}.

\section{Clothing and Social Semiotics}\label{sec:Semiotics}
Semiotics, as originally defined by Ferdinand de Saussure, is ``the science of the life of signs in society" \cite{de1985linguistic}. Semiotics investigates \textit{signs} and analyzes them to provide significance to a specific problem. There are three main elements in semiotics: the sign, what it refers to, and the people who use it. The people as social species and biological entities, instinctively evolved to survive better through facilitating living in a disciplined society by defining new signs and giving them an appropriate interpretation. Social semiotics is a subcategory of semiotics that studies how people design and interpret meanings and how these meanings are shaped by a specific social situation \cite{hodge1988social}. In social semiotics, the term \textit{resource} is preferred over the term \textit{sign} and represents a used signifier by the people to produce and to interpret communicative artefacts. In this respect, social semiotics is particularly useful in disclosing unnoticed significance and functionality of social resources and each individual is a semiotician, since everybody constantly interprets the meaning of signs around them.

Humans signify specific social context through \textit{resources} of all type, whether visual, verbal or gestural. \textit{Clothing} is a non-verbal resource that transfers meanings about individuals in the society. Cloths hold a symbolic and communicative role having the capacity to express style, identity, profession, social status, and gender or group affiliation of an individual. Although the symbolism that clothing carries on is not always clear, it evidently can be considered as the most desired personal image that one is willing to project to the society \cite{pauline2006victorian}. The study of how people use and interpret specific social context through dress is known as \textit{clothing semiotics} or \textit{fashion semiotics}, although, as stated by Sproles and Burns, clothing is distinct from fashion \cite{sproles1994changing}. In this definition, clothing is ``any covering for the human body", while fashion is ``the style of dress that is temporarily adopted by a discernible proportion of members of a social group, because that chosen style is perceived to be socially appropriate for the time and situation". Originally, clothing semiotics was studied from the fine arts perspective. Later, the perspective has been expanded and the study covered the human needs in this respect \cite{rubinstein1995dress}. Subsequently, in the 1960s, the social and psychological implications of clothing began receiving more attention from scholars. Today, clothing remains a common topic of study in social psychology \cite{owyong2009clothing} as it conveys social meaning about an individual and groups of people. It is in this way that the semiotics of clothing can be linked to the social semiotics. 

In spite of the fact that clothes have such large potential to convey a message, it must be noted that clothing semiotics understanding is complex. The social context affects the interpretation of clothing, thus, having a precise knowledge of the unconscious symbolism attached to forms, colors, textures, postures, and other expressive elements that affects the interpretation of clothing in a given culture is a desired quality in automatic analysis of this information.

\section{Clothing and Computer Vision}\label{sec:CV}

\begin{figure*}
\centering
\includegraphics[width=0.6\textwidth]{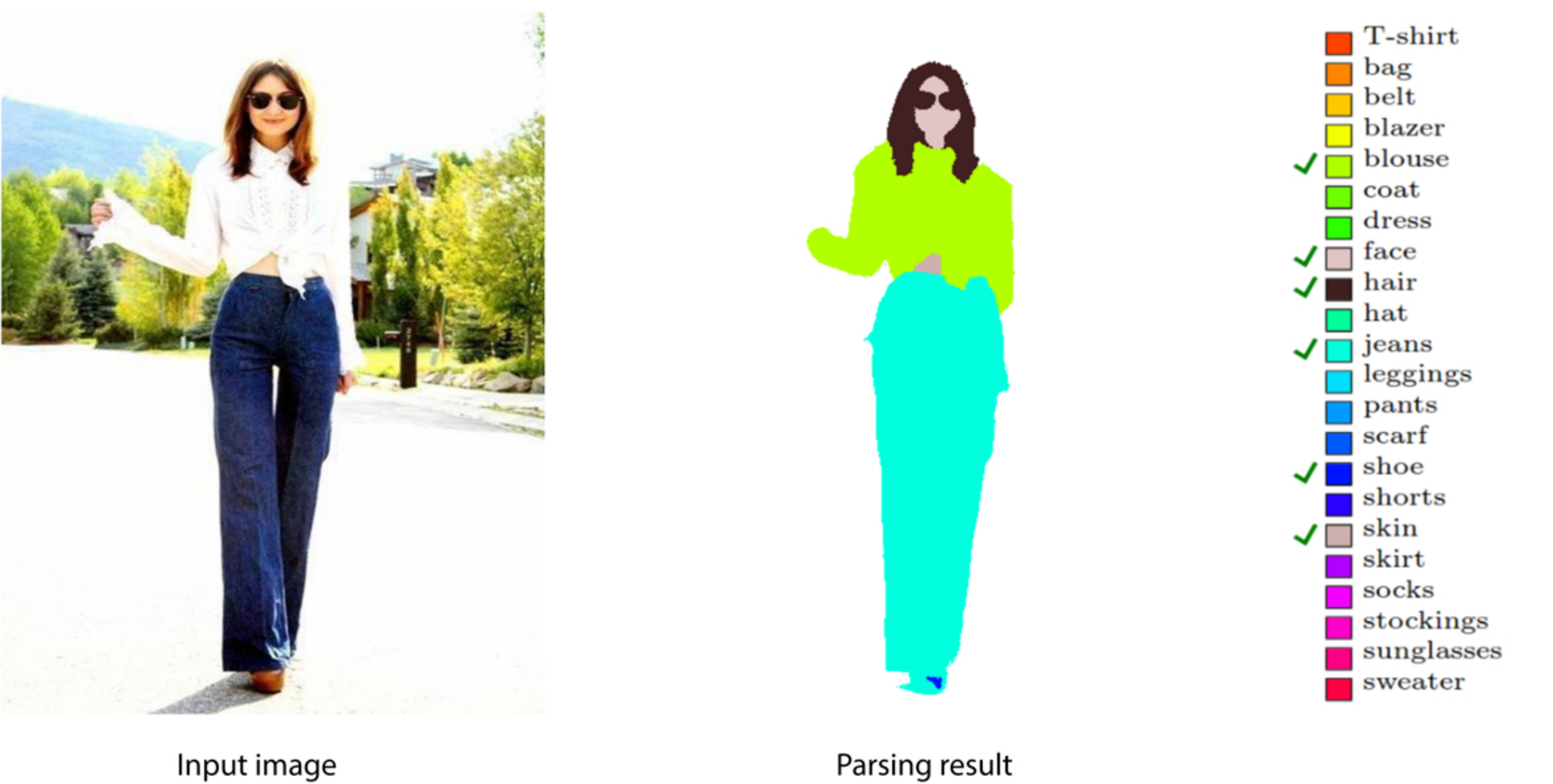}
\vspace{-0.24cm}
\caption{Parsing example. The input image (left) and the final output of parsing (right).}\label{parsing}
\vspace{-0.6cm}
\end{figure*}

\emph{Clothing style} is commonly intended in computer vision as the set of visual attributes and category labels that describe an outfit~\cite{chen2012describing, yamaguchi2015retrieving}. Examples of visual attributes are \emph{colors} (red, green, etc.), \emph{clothing patterns} (solid, striped, etc.), and more technical \emph{qualitative expressions} (skin exposure, placket presence, etc.)~\cite{chen2012describing}. 
These visual attributes have been used to measure the similarity between outfits paving the road for the identification and analysis of visual trends in fashion \cite{vittayakorn2015runway}.
The category labels are textual expressions that individuate a particular type of clothing item (shirt, sweater, etc.)~\cite{yamaguchi2015retrieving}. In most of the cases, all these textual labels are accompanied by a pixel-wise segmentation of the outfit, in which each segment is associated to a category label, and to one~\cite{liu2014fashion} or more visual attributes~\cite{di2013style}
(see Fig.~\ref{parsing}). 

This segmentation is the output of an operation commonly referred as \emph{human}~\cite{dong2014towards,liu2015matching,liang2015deep}, \emph{clothing}~\cite{yamaguchi2015retrieving} or \emph{fashion parsing}~\cite{liu2014fashion}. Clothing style can be also modeled without referring to a particular outfit, but instead to a larger set of category labels and visual attributes~\cite{kiapour2014hipster}.

Human parsing is usually performed by statistical classifiers, which operate after a training phase. The training data may consist of fully labeled data, which means images of individual outfits in which each of the pixel has a label indicating the category and/or the attribute~\cite{yamaguchi2015retrieving}. This is the most reliable source of information to train a classifier, but it is extremely cumbersome to get: in facts, manual annotation is necessary, which requires 15-60 minutes to be carried out for each single image~\cite{kohli2009robust}. Alternatively, weak labeling can be provided, which means to have training data in which an entire image is associated to a set of textual labels (in other words, the textual labels are not localized over the pixels of the outfit image). This obviously reduces the human labor to get training samples, but at the same time is less expressive, leading to classifiers which are not completely automatic: for example, \cite{liu2014fashion} requires that the testing image too comes with textual labels that indicate what to look for in the image. 

Most of the techniques for human parsing builds upon a preliminary operation, which is that of fitting a skeleton on the human body depicted in the input image. This operation is called pose estimation~\cite{yang2011articulated}, and helps to introduce a structural prior for the parsing process, which individuate salient joints (ankles, knees, hips, shoulders, elbows, wrist, neck). These points are connected by sticks forming a skeleton, which in turn drive the parsing to align with it, providing anatomically plausible segmentations~\cite{yamaguchi2015retrieving}. Unfortunately, pose estimation techniques are prone to errors in the case of missing data, due to occlusions or auto-occlusions; for this reason, images of single persons where the entire body is portrayed, are preferred. Images depicting parts of the body (as those ones captured via wearable sensors, where usually the whole body does not fit) represent a serious issue. In addition, pose estimation is weak in the case of large and long clothing, covering the structure of the body for what concerns some of the joints (a person wearing long dress has its knees completely covered). This issue has been recently faced by facing human parsing and pose estimation as two intertwined aspects of the same problem, introducing the concept of \emph{semantic part} (such as leg, arm, head)~\cite{dong2014towards}. A semantic part can be iteratively modeled with tools usually employed for human parsing (as the \emph{Parselets}~\cite{dong2016parsing}) and as an ensemble of joints, taking from the pose estimation literature.

\section{Clothing style and social signals: the social signal processing point of view}\label{sec:Our}

Social signal processing (SSP) ``aims at providing computers with the ability to sense and understand human social signals''~\cite{vinciarelli2009social}.
To this aim, Vinciarelli et al. \cite{vinciarelli2009social} suggest a four-step pipeline in which, after having recorded the scene and detected humans (step 1 and 2), in the step 3, feature extraction has to be performed, where features are behavioral cues whose interpretation brings to individuate social signals. In the step 4, social signals have to be grounded with the scene context, in order to understand social interactions. We are interested specifically in the last two steps of the pipeline, since we assume that the scene has been already recorded and the individuals have been properly detected.  

In the rest of this section, we will individuate the research questions (indicated with the letter \textbf{Q}) that can be inserted in these two steps, providing our intuitions about possible answers (letter \textbf{A}), driven by the literature of the human sciences and/or our speculations, together with the type of experiments we would like to carry out, to provide the community with deeper insight and novel tools for clothing social signaling.


\subsection{Clothing behavioral cues as an individual social signal}

\emph{A}-\textbf{Q1 - How much clothing-related cues are independent from other standard behavioral cues, in the determination of particular social signals?}
The question essentially asks how the mapping from visual features related to clothing (for example, the type and appearance of a particular clothing item, e.g., a shirt) has to be carried out in dependence from other cues such as the ones reported in~\cite{vinciarelli2009social} (Table~1,~pag.1745). In other words, this question is very preliminary and asks for a feasible and reliable protocol with which clothing-related cues can be analyzed without caring of the effects due to other features in determining social signals.

\begin{figure}
\begin{center}
\includegraphics[height=0.2\textwidth]{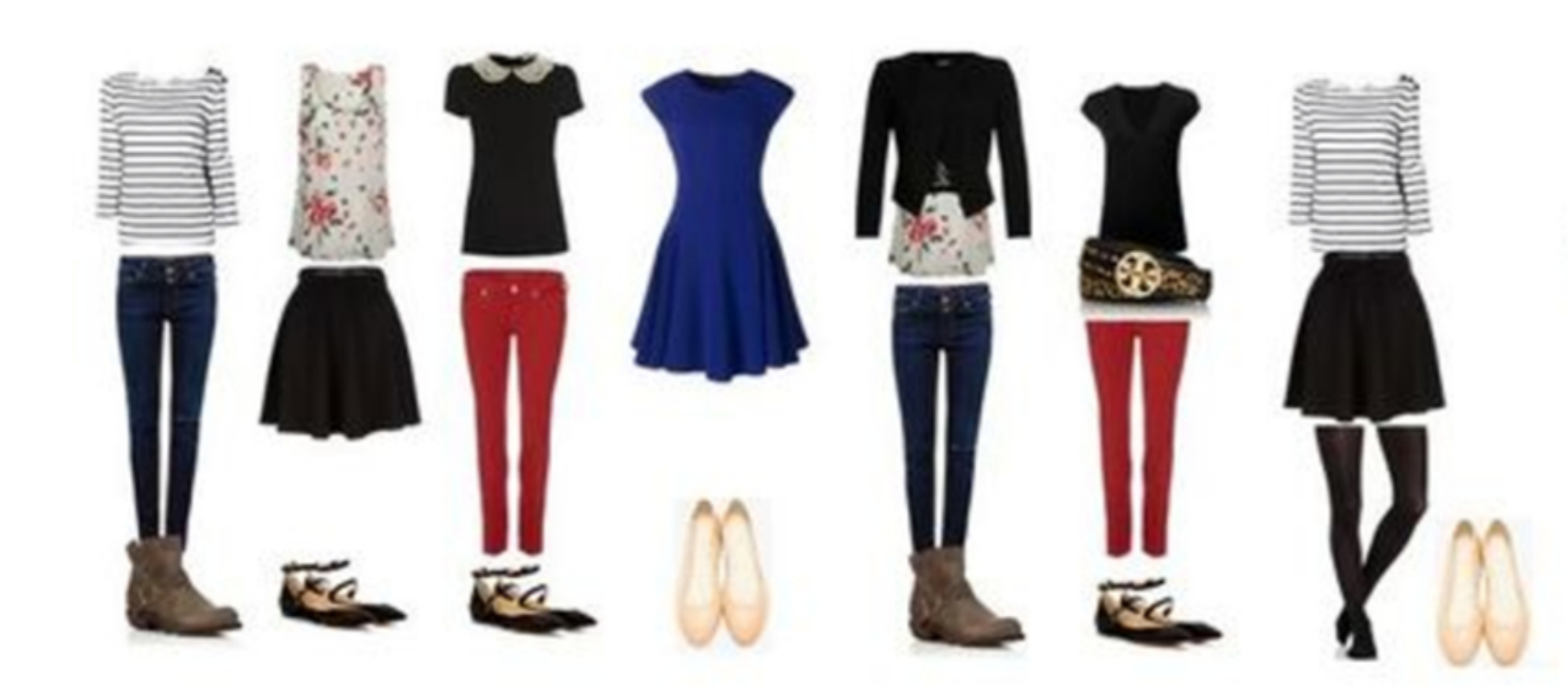}
\vspace{-0.24cm}
\caption{The picture shows an example of clothing outfits typical of online shops. }\label{outfit}
\vspace{-0.6cm}
\end{center}
\end{figure}

\emph{A}-\textbf{A1 -} In social situations a clothing outfit comes with the body that wears it, so that other cues, in particular related to physical appearances, gesture and posture, face, emotional expressions and eyes behavior ~\cite{vinciarelli2009social, frith2009role} are obviously co-present and some cues may have different effects depending on the visible human body. For example, facial expression comes more into vision if only the upper body is visible. This could be the reason why online shops often present garments without the human body (Fig. \ref{outfit}). Thus, an analysis on these data seems reasonable and may help for answering A-Q1.

A simple yet important experiment would be that of checking whether the presence of different types of body appearances will change the nature of the social signal transmitted. In particular, our first step is to enrich the annotation of a clothing dataset, for example, the \emph{Exact Street2Shop} dataset \cite{hadi2015buy}. For a given garment, the dataset contains some ``shop pictures", where the garment is usually located on a neutral background, without being worn by a human body. Together with this, the dataset offers a ``street picture", where the same garment is worn by a subject among an undefined set of people. The idea is to first annotate standard semantic information about thepeople in the street photos (gender, expressions etc.). Next, different assessors will evaluate the street and the shop photos, defining the person wearing that particular garment in terms of social signals and personality traits. In the case of the street photos, the person in the picture is present and annotated, while in the case of the shop photos, persons are absent.
The goal is to discover whether the presence of the person changes significantly the judgment of the assessors, and if this correlates with the semantic information associated to the person.

A finer setup, given a particular person, could be that of isolating the most the cues related to clothing by masking behavioral cues coming from the face (blurring the face) or hiding the height (removing the background scene). The interrelation between behavioral cues and other features in terms of social signals has never been investigated in the literature.

\begin{figure}
\begin{center}
\includegraphics[width=0.5\textwidth]{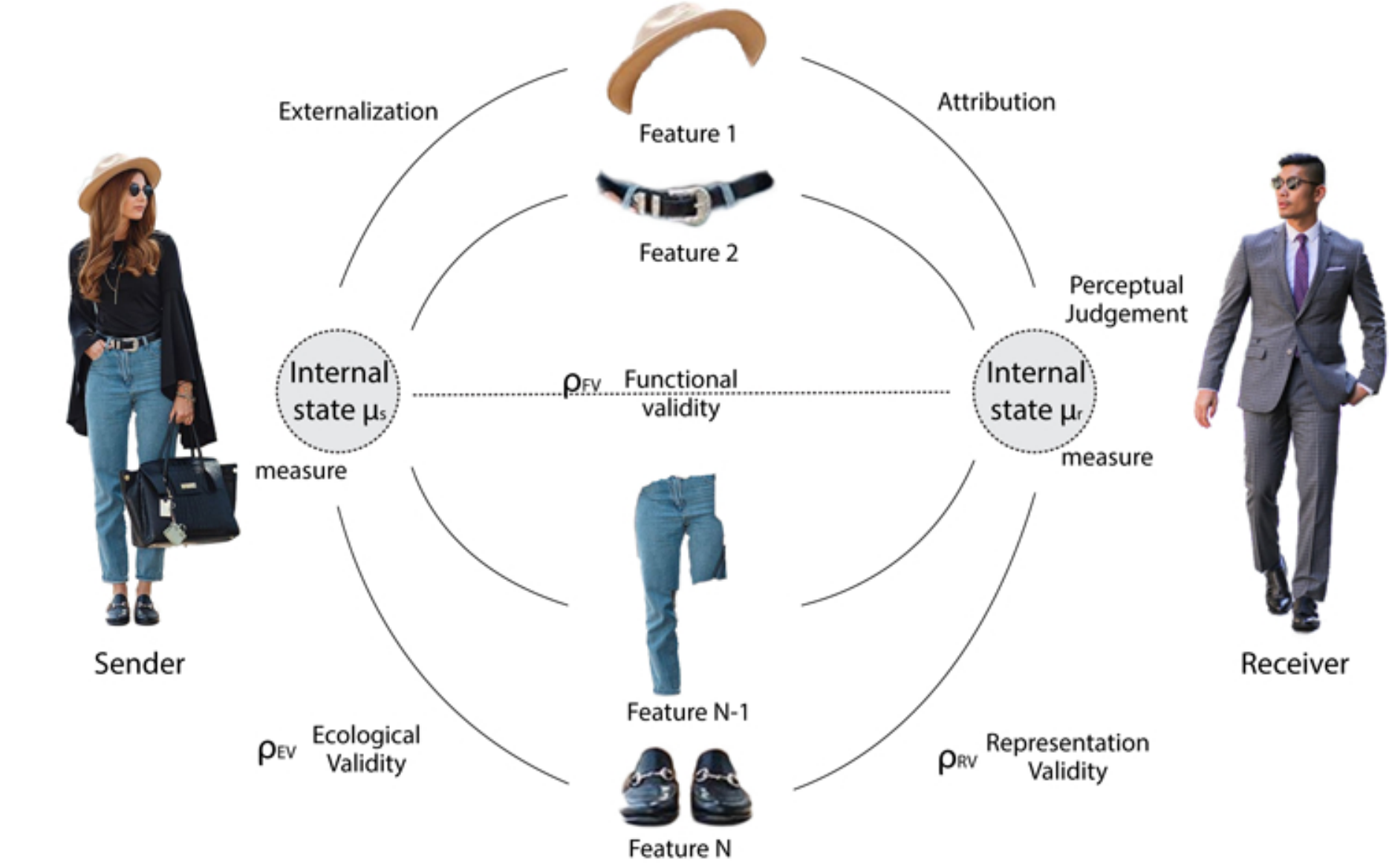}
\vspace{-0.24cm}
\caption{The picture shows a simplified version of the Brunswik Lens Model adapted to the transmission of a social signal between a Sender and a Receiver.}\label{brunswik}
\vspace{-0.6cm}
\end{center}
\end{figure}

\emph{A}-\textbf{Q2 -  How to evaluate the nature of a social signal generated by clothing behavioral cues?} For understanding this question, one may consider the \emph{Brunswick's Lens} model. A simplified version of this model is adopted by Cristani et al. ~\cite{cristani2013unveiling} (see Fig.~\ref{brunswik}). In few words, the model says that a social signal is not necessarily univocally intended. More in the detail, the model assumes that a social signal is sent by a sender, $S$, as a consequence of its internal state, $\mu_s$, which is assumed to be measurable. For example, $S$ feels himself extrovert (his internal state), and this awareness is measured by a self-assessment (for example, using the Big Five questionnaire~\cite{caprara1993big}). $S$ wears some clothing items and as we are assuming that clothing items are related in some way with the internal state of $S$, they can be assumed as an \emph{externalization} of the internal state. The receiver $R$ sees the clothing items worn by $S$, and infers about the internal state of $S$, which in this case is  $\mu_r$, to highlight that possibly is not equal to $\mu_s$. This process, called \emph{attribution}, which brings to a perceived state what can be measured itself. The Brunswick Lens model states that a social signal has high ecological validity $\rho_{\text{EV}}$, if there is a high correlation between the internal state of $S$ and the features. Viceversa, a social signal has high representational validity $\rho_{\text{RV}}$, if the correlation between the features and the state inferred by $R$ is high. Finally, if the internal state of $S$ correlates with the inferred state of $R$, it means that the communication through the social signal mediated by the features has high functional validity $\rho_{\text{FV}}$. 

\emph{A}-\textbf{A2 -} The Brunswick's Lens essentially states that the nature of a social signal should be measured considering the sender of the signal and the receiver. This opens up to diverse experiments, suggesting a protocol for each one of them. For example, in order to understand how a particular social signal built upon clothing behavioral cues is interpreted by a generic receiver, it is necessary to measure the perceived state of multiple assessors. If the correlation between the behavioral cues and the perceived state of the assessors is high, we may individuate the \textit{implied} meaning of a particular outfit. In more practical terms, to assess whether an athletic outfit is a behavioral cue that communicates the social signal of extroversion, this can be asked to a set of assessors. If the features that characterize the athletic outfit correlate with the extroversion assessment, then this message can be understood that athletic outfit triggers a certain reaction in a generic audience in terms of social signals. In order to individuate the attributes that most consistently originate social signals, deep learning technologies will come into play. One of the most attractive features of deep architectures is that they can be ``opened'' and ``visualized'', allowing to easily interpret what is codified into the internal layers \cite{zeiler2014visualizing,mahendran2015understanding,yosinski2015understanding,Evaluating2016SamekTNN}. Exploiting these strategies, once annotations of social signals have been extracted from garment images, the goal would be that of feeding them into deep architectures, capturing the most discriminative visual patterns. In this way, the generic semantic label of ``athletic outfit'' can be explained in terms of behavioral cues (in the sense of~\cite{vinciarelli2009social}), like shape, color and texture attributes. 



\emph{A}-\textbf{Q3 - Is there an agreement between one's self-image and the impression conveyed to others through his/her clothing style?}
It has long been known that \textit{clothing affects how other people perceive us as well as how we think about ourselves}. This question asks whether there is a consistency between self-perception of an individual and perception of other people towards him/her.

\emph{A}-\textbf{A3 -} Often people choose what they wear as a means of self-expression. The individual measurement of the effect of clothing on self-perception and perception of others, has been studied previously by Heart \cite{heart2011apparel}. However, the question of whether others perceive the desired message that the person wishes to communicate, has not been explicitly studied before. An experimental set up should first facilitate separate investigation of clothing effect over self-perception of an individual and perception of others over them and then study their correlation.


\emph{A}-\textbf{Q4 - Which clothing behavioral cues are related to the social signal of the \emph{attractiveness}?}
This question asks if clothing style influences the perception of attractiveness of others. Attractiveness is the main social signal associated to physical appearance~\cite{vinciarelli2009social} and attractive people tend to be considered as having high status, good personality, and being extrovert.  

\emph{A}-\textbf{A4 -} 
Clothing is considered as an indicator of socioeconomic status~\cite{sybers1962sociological,kerr1990dress} and personality traits~\cite{slepian2015cognitive}. According to Johnson et al. \cite{johnson2008dress}, the most investigated concepts using dress manipulations were \emph{dress}, \emph{status}, and \emph{attractiveness} and listed the most  experimented dress manipulation to study the effect of clothing on attractiveness concept as grooming, tidiness, makeup, and, natural physical appearance such as hair and eye color, height and, weight (see \cite{johnson2008dress}, Table 3). Clothing can be strongly related to arise perception of attractiveness in people towards a person. To prove this hypothesis, a possible experiment would be that of capturing the influence of single clothing items, or multiple clothing elements arranged in an outfit towards the definition of an attractive person. In the same line with A-Q1, other behavioural cues should be considered, selected or masked, so to avoid explaining away effects. Also in this case, deep learning architectures and tools to visualize them (\cite{zeiler2014visualizing,mahendran2015understanding,yosinski2015understanding,Evaluating2016SamekTNN}, see A-A2) will help in segregate and explicitly individuate visual attributes that convey the impression of attractiveness.


\emph{A}-\textbf{Q5 - How much impact clothing have on the other individual’s first impression?} The importance of first impression comes more into attention in relation to its effect on the overall lasting impression. The last impression is what we will remember most about a social situation, however, one probably will not have a last impression if do not get the right first impression.

\emph{A}-\textbf{A5 -} ``You never get a second chance to make a first impression" as noted by Oscar Wilde. Although a large amount of cues aggregate together to form a first impression, we hypothesise clothing is a strong cue that eases the process for the people \cite{barnard2002fashion} to make a first impression. This quick judgment that happens in less than a minute \cite{todorov2009evaluating}, can lead us towards a set of assumptions about a set of personal traits of that person, such as attractiveness, likability, competence, and aggression \cite{willis2006first}. Howlett et al. \cite{howlett2013influence} studied the effect of clothing alone on the first impression and reported that clothing solely influences the first impression of the others even in limited exposure time. To detect the influence of individual’s first impressions, the labeling of the \emph{Exact Street2Shop} dataset (see A-A1) can be performed including the time dimension into play, enforcing the user to give a quick answer on the impression the clothing does trigger, explaining then by textual attributes the item(s) that leads him to such an answer.



\subsection{Grounding clothing related social signals with scene context}

\emph{B}-\textbf{Q1 - How much clothing-related cues help in capturing the context of a social interaction?} The idea here is to study how clothing items worn by people involved in a focused or unfocused interaction \cite{setti2015f} can tell about the interaction itself.

\emph{B}-\textbf{A1} - People wearing outfits of a very similar kind, different from that of the rest of the crowd, are connected with a high chance, and this in turn helps in individuating the nature of a social interaction. Sport players with the same attire and supporters with the same t-shirts in a spectator crowd could be considered as an example of this connection. In this case, simple counting algorithms, specialized to finding similar items in a scene, may be of a great help~\cite{SettiCount2017}. However, the problem becomes more challenging when it comes to other types of interactions, namely ordinary exchanges in generic scenarios (waiting in a bus stop, attending a conference, etc.). An ideal solution is to develop models capable of first, assigning clothing visual attributes to social scenarios (learning the most expected garments on the beach, in a Starbucks, during a conference etc.), and second, individuating similarities among outfits, from those very explicit (team outfits that are different only for the number depicted on the shoulders) to those more insidious to catch (individuating people that bring the same bag to individuate a social event). The interplay with social sciences lies in motivating the results from the learning stage of the model, that is, analyzing and interpreting the most emblematic garments for a particular interaction. Even in this case, deep learning technology will help, especially those architectures equipped with region proposal modules \cite{ren2015faster}. The novel idea here could be that of assuming the region proposal module as on-line evolving, drifting towards the detection of people exhibiting similar clothing. 


\emph{B}-\textbf{Q2 - How the clothing style drive people to socialize? Specifically, do people with the certain type of apparel socialize with similarly suited people?} This question asks whether our higher tendency to socialize with certain people is influenced by the clothing they wear, and if people tend to socialize more with similarly dressed people.

\emph{B}-\textbf{A2 -} The approach towards answering this question is twofold. On one side, clothing in the same way as being considered as a flag to make visible a specific ideology, culture, or ethnicity, it also can be considered as a social catalysts among similarly dressed people. In an example, Nash \cite{nash1977decoding} studied the influence of dressing on runners and stated that when two runners are dressed alike they engaged in an extended conversation as opposed to a short nonverbal greeting that occurred among runners that dressed differently from each other.
On the other side, the effect of clothing on the people's self-perception, leads to variations in their social relations. As an example, feeling comfortable is an important factor in a social interaction and clothing has the power to make a person feel comfortable or not. Simply, when clothing can be used as a criterion for judgment, people may unconsciously feel judged and act according to it. 
The connection with pattern recognition here lies in the approaches for detecting gatherings of people, which are proven to be very robust and versatile to diverse types of scenarios~\cite{setti2015f,shao2014scene}. Applying pattern recognition techniques for correlating clothing types of interactants will unveil possible affinities which may facilitate social interactions.

\section{Concluding remarks}\label{sec:Conc}

This work presents some research questions that are related to the investigation of the social signals associated to clothing. The outcome of these questions, other than filling a gap in the social signal processing literature, may have important relapses. For example, it could facilitate the design of online personal stylists able of indicating, on the one side, the type of impressions one's outfit may trigger (associated to their particular body or posture), and on the other side, which are the most suitable outfits for attracting the attention of others. This has helpful applications in facilitating social interactions. 
\bibliographystyle{abbrv}
\bibliography {Bibliography}

\end{document}